\documentclass[conference]{IEEEtran}
\IEEEoverridecommandlockouts

\usepackage{cite}
\usepackage{amsmath,amssymb,amsfonts}
\usepackage{algorithmic}
\usepackage{graphicx}
\usepackage{textcomp}
\usepackage{xcolor}
\usepackage{comment}
\usepackage{soul}
\usepackage{url}
\def\BibTeX{{\rm B\kern-.05em{\sc i\kern-.025em b}\kern-.08em
    T\kern-.1667em\lower.7ex\hbox{E}\kern-.125emX}}
\begin{document}

\title{Rapidly Built Medical Crash Cart! Lessons Learned and Impacts on High-Stakes Team Collaboration in the Emergency Room}

\author{\IEEEauthorblockN{Angelique Taylor}
\IEEEauthorblockA{\textit{Information Science}\\
\textit{Cornell University }\\
New York, USA \\
amt298@cornell.edu}
\and
\IEEEauthorblockN{Tauhid Tanjim}
\IEEEauthorblockA{\textit{Information Science}\\
\textit{Cornell University }\\
New York, USA \\
tt485@cornell.edu}
\and
\IEEEauthorblockN{Michael Joseph Sack}
\IEEEauthorblockA{\textit{Information Science}\\
\textit{Cornell University}\\
New York, USA \\
mjs596@cornell.edu}
\and
\IEEEauthorblockN{Maia Hirsch}
\IEEEauthorblockA{\textit{Mechanical Engineering}\\
\textit{Israel Institute of Technology}\\
Haifa, Israel \\
mh2567@cornell.edu}
\and
\IEEEauthorblockN{Kexin Cheng}
\IEEEauthorblockA{\textit{Information Science}\\
\textit{Cornell University}\\
New York, USA \\
kc2248@cornell.edu}
\and
\IEEEauthorblockN{Kevin Ching}
\IEEEauthorblockA{\textit{Emergency Medicine} \\
\textit{Weill Cornell Medicine}\\
New York, USA \\
kec9012@cornell.edu}
\and
\IEEEauthorblockN{Jonathan St. George}
\IEEEauthorblockA{\textit{Emergency Medicine} \\
\textit{Weill Cornell Medicine}\\
New York, USA \\
jos7007@cornell.edu}
\and
\IEEEauthorblockN{Thijs Roumen}
\IEEEauthorblockA{\textit{Information Science}\\
\textit{Cornell University}\\
New York, USA \\
thijs.roumen@cornell.edu}
\and
\IEEEauthorblockN{Malte F. Jung}
\IEEEauthorblockA{\textit{Information Science}\\
\textit{Cornell University}\\
New York, USA \\
mfj28@cornell.edu}
\and
\IEEEauthorblockN{Hee Rin Lee}
\IEEEauthorblockA{\textit{Media \& Information} \\
\textit{Michigan State University}\\
East Lansing, Michigan \\
heerin@msu.edu}
}

\maketitle

\begin{abstract}
Designing robots to support high-stakes teamwork in emergency settings presents unique challenges, including seamless integration into fast-paced environments, facilitating effective communication among team members, and adapting to rapidly changing situations. 
While teleoperated robots have been successfully used in high-stakes domains such as firefighting and space exploration, autonomous robots that aid high-stakes teamwork remain underexplored. To address this gap, we conducted a rapid prototyping process to develop a series of seemingly autonomous robot designed to assist clinical teams in the Emergency Room. 
We transformed a standard crash cart—which stores medical equipment and emergency supplies into a medical robotic crash cart (MCCR). 
The MCCR was evaluated through field deployments to assess its impact on team workload and usability, identified taxonomies of failure, and refined the MCCR in collaboration with healthcare professionals. 
Our work advances the understanding of robot design for high-stakes, time-sensitive settings, providing insights into useful MCCR capabilities and considerations for effective human-robot collaboration. 
By publicly disseminating our MCCR tutorial, we hope to encourage HRI researchers to explore the design of robots for high-stakes teamwork.
\end{abstract}



\section{Introduction}
 

\begin{figure}[t]
    \centering
 \includegraphics[width=0.35\textwidth]{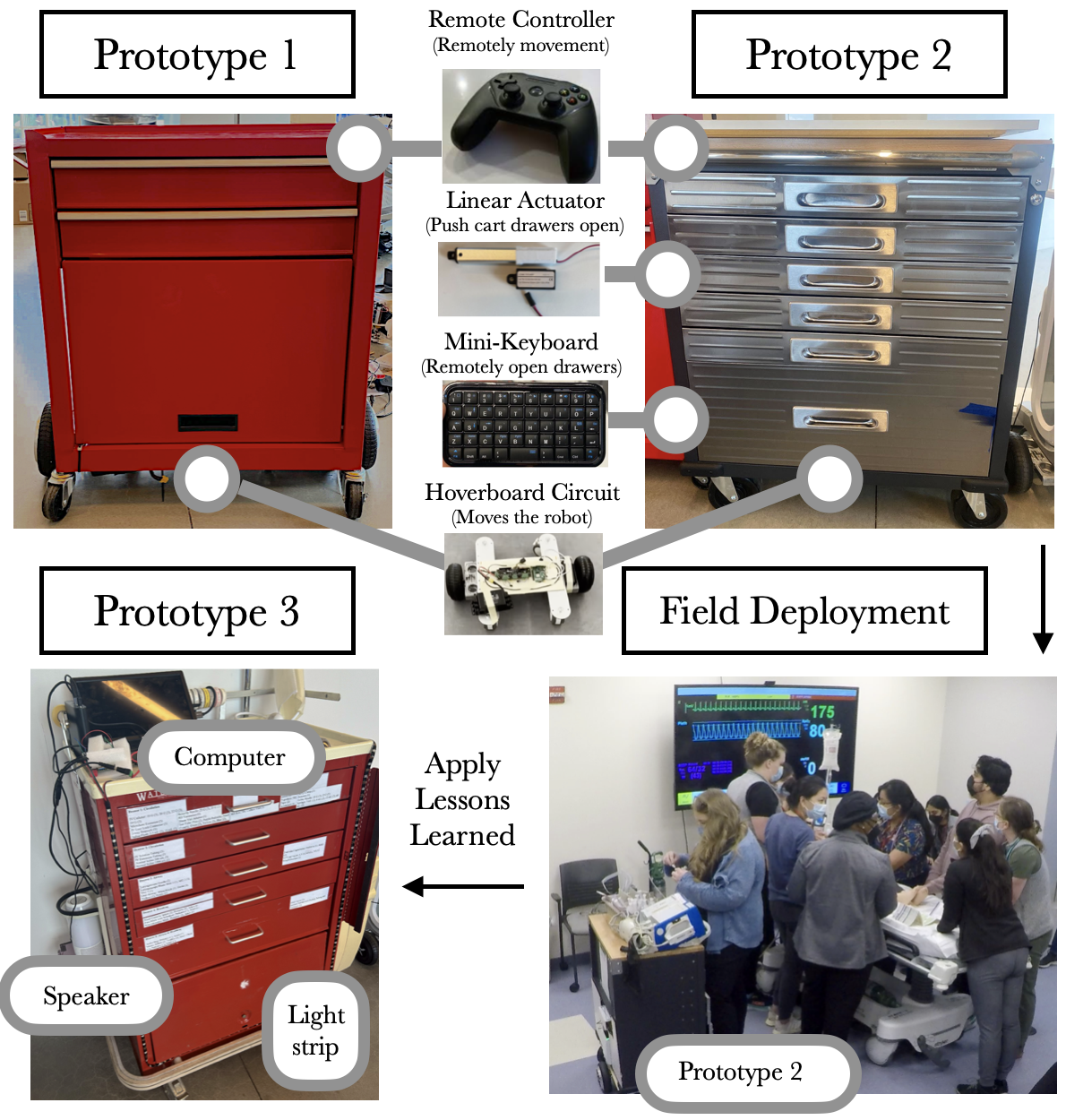}
    \caption{We built three teleoperated medical crash cart robots (MCCRs). MCCR 1 delivers supplies using a hoverboard circuit. MCCR 2 delivers supplies, recommends supplies using drawer opening capabilities, and was deployed at a medical training event which revealed insights. This led to the MCCR 3 design which recommends supplies and generates tasks reminders using drawer lights, speech, and alerts.}
    \label{fig:cart_cart_field_studies}
    \vspace{-5pt}
\end{figure}


Teleoperated robots have become indispensable tools for action teams---highly skilled specialist teams that collaborate in short, high-pressure events, requiring improvisation in unpredictable situations \cite{sundstrom1990work}. 
For example, disaster response teams rely on teleoperated robots and drones to aid search and rescue operations \cite{casper2003human, murphy2017disaster}. 
High-stakes military and SWAT teams use teleoperated ordnance disposal \cite{carpenter2016culture} and surveillance robots \cite{jones2002autonomous} to keep the teams safe. 
Surgical teams employ teleoperated robots to perform keyhole surgeries with a level of precision that would be unimaginable without these machines \cite{cheatle2019sensing, beane2019shadow}. 
Despite these advances, current robotic systems for high-stakes teams remain teleoperated, and questions about how to increase their autonomy for such teams remain underexplored.

A growing body of human-robot interaction (HRI) research examines the design of autonomous robots to better support teamwork. 
For example, recent work has explored how robots can contribute to group conversational dynamics and collaborative learning processes \cite{tennent2019micbot,alves2019empathic}. 
Other work has addressed teamwork, which involves activities such as action teams \cite{jamshad2024taking}, navigation, and lifting objects \cite{fourie2022joint}. 
Despite a few notable exceptions (e.g. \cite{jamshad2024taking,bethel2012discoveries, jones2002extreme}), prior work has focused on low-stakes teamwork such as problem-solving or decision-making teams. 
Designing autonomous robots to support high-stakes teamwork in emergency settings presents unique challenges that our work aims to explore. 
This includes seamless integration into fast-paced settings, facilitating effective team communication, and adapting to rapidly changing situations.

We present the design, development, and deployment of seemingly autonomous medical crash cart robots (MCCRs) into action teams. 
Our work focuses on high-stakes teamwork in the emergency room (ER) which presents unique challenges including time pressure, specialized expertise, and unique demands for communication. 
The ER also presents challenges for the integration of autonomous robots: loud noises from sensors, alerts, team communication, and visual occlusions through tightly crammed ER equipment make perception challenging. 
Tight spaces provide little room for navigation and the nature of the work leaves no room for error.

By leveraging the concept of embodiment \cite{massaro1990psychology}, our approach integrates robots into medical crash carts or 'code carts', a tool used to store medical supplies and equipment. This form factor is well-suited to provide multimodal feedback to support team decision-making during medical procedures. 
We highlight key findings 
including the identification of useful capabilities for robots working with teams in time-sensitive, high-stakes settings, and encourage others to explore the design of robots from stakeholders' perspectives.


Our research \textbf{contributes} 1) knowledge about the iterative design process of building new robots that engage with users, lessons learned throughout these iterations, and HCWs' perspectives of robots in safety-critical high-stakes environments and 2) release a publicly available robotic development tutorial and toolkit for the ER including Github code and documentation of circuit diagram, electrical components, and supplies\footnote{\url{https://github.com/Cornell-Tech-AIRLab/crash\_cart\_robot\_tutorial}}.



\section{Related Work}

\subsection{Human-Robot Collaboration} 

The field of human-robot interaction (HRI) has long studied effective ways for robots to engage in collaborations with humans. 
Prior work human-robot collaboration in terms of conversational dynamics and physical human-robot collaboration.
For example, prior work has explored how robots can shape conversation dynamics in group collaboration \cite{tennent2019micbot} and how robots can improve group learning processes \cite{alves2019empathic}. 
Furthermore, physical human-teaming involves physical activities such as moving around, lifting items, moving items from one place to another. 
The literature often frames this as a joint action, adaptation, and entrainment problem which models psychological, neurological, and physical mechanisms by which humans collaborate with robots \cite{fourie2022joint,iqbal2016movement,iqbal2017coordination}. 
Furthermore, prior work often involves evaluating robots in well-controlled environments with minimal consequences for their actions, whereas in acute care settings which are high-risk environments, human actions could result in patient safety risks. 
HRI in \textit{action teams} is the most relevant work as it highlights the importance of well-designed proactive robot behaviors to address operational failures in time-critical contexts (i.e., healthcare and firefighting) \cite{jamshad2024taking}. 
Prior research has explored modeling techniques for human intent, human collaborations with robots, and methods that enable robots to anticipate human actions \cite{levine2014concurrent,nikolaidis2013human,dominey2008anticipation,hoffman2007effects}.
As a result, further research is required to understand how robots can assist in team collaborations in ER environments.

\subsection{Collaboration in Medical Teams}

There are many robots designed to support people in terms of health and wellbeing \cite{kyrarini2021survey,riek2017healthcare}.
For example, assistive robots are used as companions to support older adults \cite{yang2017companion}, robotic wheelchairs are used to support patient mobility \cite{ktistakis2017assistive,jiang2016enhanced}. 
Robots are used to support people with rehabilitative training, patients with psychiatric disabilities by engaging in rehabilitative training \cite{sato2020rehabilitation}, older adults to support recreation  \cite{carros2020exploring}, and improve patients' motor skills \cite{wu2016design}.
Robots also perform non-patient-facing tasks, such as fetching and delivering supplies \cite{ahn2015healthcare,taylor2019coordinating,taylor2021social,taylor2020situating,taylor2021human} to free up time for HCWs to focus on patient care. 
They are also used to support nurses with triage \cite{ahn2015healthcare}, lifting patients \cite{lee2014design}, and telemedicine \cite{matsumoto2023robot}.

\subsection{Embodiment of Robots}

Much prior research in HRI demonstrates that robot embodiment sets the expectations of how robots can interact with people based on robot affordances \cite{norman1999affordance,gibson20133}, shapes how people perceive robots \cite{passman1975mothers,rani2004anxiety, bauer2008human}, and to what extent people adopt robots \cite{riek2017healthcare}.
For example, robots come in many shapes including humanoids \cite{bauer2008human,hayashi2007humanoid,kose2009effects,krogsager2014backchannel}, zoomorphic (animal-like) \cite{short2017understanding,kalegina2018characterizing}, ottomans \cite{sirkin2015mechanical}, and even adjustable wall robots \cite{nguyen2021exploring,wang2019design}, and adjustable furniture \cite{hauser2020roombots,nigolian2017self}.
Robots can also vary in terms of how human-like or machine-like they appear where those that appear too human-like often appear uncanny to humans \cite{strait2015too,koschate2016overcoming} and those that appear more machine-like are often viewed as companions or pets \cite{loffler2020uncanny}. 
Most similar to our work is the work done by Ju et al. \cite{mok2015place} who designed an automatic drawer open mechanism for robot carts in office spaces.
Building on this work, we focus on building robots for safety-critical healthcare settings and we focus on building robots for team-based interactions, opposed to dyadic interactions. Taylor et. al.'s work, \cite{taylor2019coordinating,taylor2024towards} motivates the use of crash cart form factors as a way to integrate robots into clinical team collaborations. A recent study revealed robotic use-cases in clinical team settings which we build on in our work to conduct iterative rapid prototyping of new MCCRs.

\subsection{Design Approaches for Field Studies}

Participatory design (PD) has gained increasing popularity in HRI as a way to invite stakeholders to act as co-designers \cite{lee2017steps, azenkot2016enabling, antony2023co}. 
Despite this growing interest, PD has been used limitedly to generate initial design ideas. 
However, PD researchers outside the HRI community have emphasized the importance of long-term ongoing efforts, advocating for the PD process as a form of `infrastructuring' \cite{bjorgvinsson2010participatory}. 
This concept highlights that prototypes (or existing technologies) are merely entry points situated within the complex networks of the communities \cite{suchman2002located, star1994steps}. 
Inspired by the concept of infrastructuring, we have focused on benefiting healthcare teams through a long-term collaboration (since 2021) with a healthcare professionals’ organization that conducts an annual bootcamp for interprofessional emergency medicine team training. 
Our work involved an entangled process of designing, developing, deploying, and redesigning our MCCRs to ensure they align well with the needs of our HCWs. This study will demonstrate understudied aspects of PD that view the design process as a continuous dynamic endeavor. 
This new approach is particularly beneficial to the HRI community, considering the importance of prototyping incorporate unique factors that differ depending on the setting \cite{lee2022configuring}.


%
\section{Methodology}
          

Over the past several years, we collaborated with two medical educators in Emergency Medicine, with 18 and 20 years of experience respectively in medical education and practice. Their wealth of experience in medical education is well-suited for our iterative feedback loops to collect design requirements of MCCRs that can enhance clinical workflows. 
Robots are not found in emergency room (ER) settings that assist team collaborations during medical procedures introducing the goal of creating a wizard of OZ platform for healthcare workers to control and provide ideal behavior. 
As already mentioned, the cost of mistakes is astronomical (to the point of life-threatening consequences) making the risk of exploring full automation unacceptable at this point. 
In order to fulfill and enact autonomous decision making we seek to understand the types of feedback the robot should provide, and the impact of these capabilities on team dynamics.
Throughout this rapid design process, we collected feedback on the robot's capabilities, potential use-cases, and concerns.

\subsection{Design Considerations}

We present key MCCR design factors: appearance, navigation, supply recommendations, and decision-support.

\textbf{Appearance:} Our rapid prototyping process begins with a cart that does not resemble the appearance of a traditional crash cart to further our understanding of appropriate navigation capabilities. 
However, feedback from our medical collaborators emphasized the importance of making the MCCR easily identifiable as a crash cart to ensure effective supply retrieval. 
MCCRs that do not resemble the crash cart could cause confusion and a lack of adoption when stakeholders search for carts to treat patients. 
We redesigned the prototype to increase their resemblance to traditional crash carts throughout our design process, prioritizing clarity and adoption.

\textbf{Navigation:} By leveraging the benefits of traditional crash carts in ERs, we sought to design new MCCR capabilities, including a navigation system to enable the crash cart to deliver relevant supplies to healthcare workers.
Patient rooms are tight spaces, making it unclear how a robot could appropriately navigate to users while avoiding occlusions, and disrupting patient care.
Thus, we started by designing a teleoperated MCCR platform that can allow us to collect expert observations of appropriate robot behaviors from stakeholders. 

\textbf{Supply Recommendations:} Another factor introduced in our design process was a MCCR supply recommendation system. 
HCWs often shuffle through drawers to locate relevant supplies and medication which could lead to care delays. 
As a result, we explored ways for the MCCR prototypes to communicate supply recommendations to streamline item retrieval processes and reduce care delays.

\textbf{Decision-Support:} MCCRs are well-suited to provide decision-support to healthcare workers because many medical errors originate from the item retrieval process. 
For example, due to the loud noises in the patient room and fast-paced nature of care tasks, healthcare workers may retrieve incorrect items (e.g. wrong medication dosage) which can lead to patient safety risks. 
Thus, our rapid prototyping process consisted of designing capabilities for the MCCR to communicate with healthcare workers to assist with decision-support in terms of indicating which and when errors occur.

 \begin{figure}[t] 
	\centering 
	\includegraphics[width=1.0\linewidth]{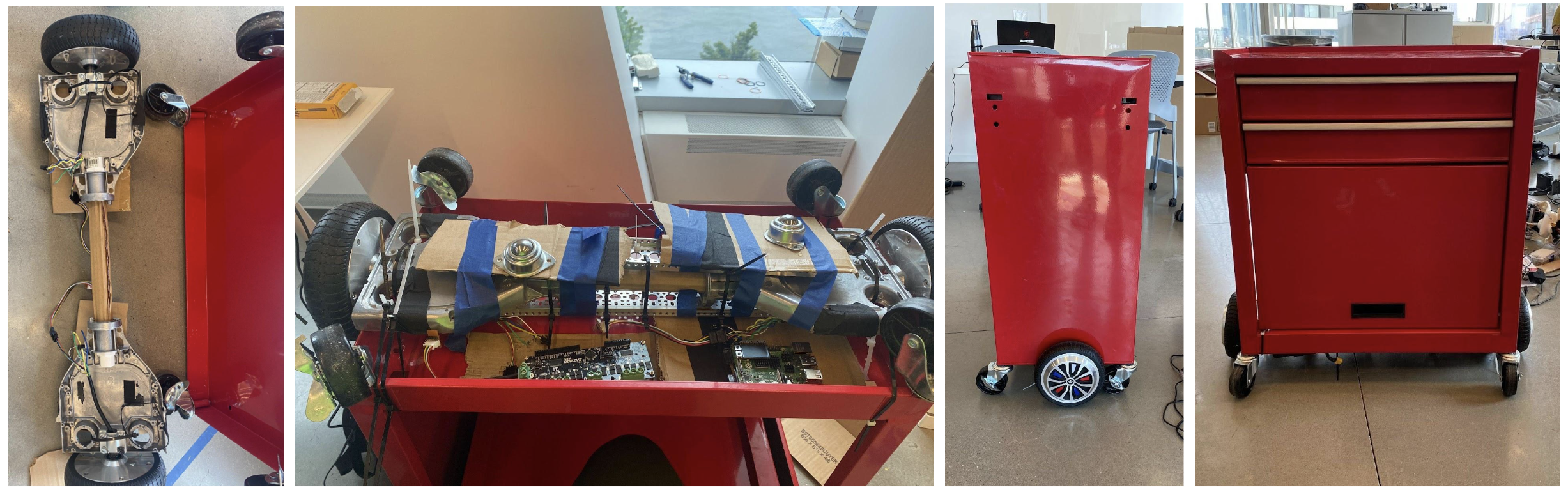} 
	\caption{We built Prototype 1 by connecting the Garbatrage hoverboard circuit \cite{mandel2023recapturing} to a tool cart using metal chassis, cardboard, and tape.} 
	\label{fig:robot_version1} 
\end{figure}

\subsection{Crash Cart Robot Prototype 1}

The objective of this prototype was to be flexible and modular to serve as a boundary object for communication with different stakeholders. 
Its core functionality is centered around mobility \textcolor{black}{(}see Figure \ref{fig:cart_cart_field_studies}). Each function was accomplished by effectively slapping together off-the-shelf modules on a standard workshop shelf.
We chose this cart as the first prototype because it resembles a crash cart in terms of its red color but all the drawers do not open in a similar fashion as a medical crash cart.
Nevertheless, these prototypes provided an opportunity for us to rapidly add mobility to cart-base objects and explore how robots could navigate in ERs.

We built the first MCCR prototype built upon the Garbatrage framework \cite{mandel2023recapturing} as a motorized base for a red tool cart that was being disposed of (see Figure \ref{fig:robot_version1}). 
The “Garbatrage” framework is a hardware platform built on a hoverboard, hands-free motorized scooter.
This platform offers many benefits including repurposed hardware components and devices, recaptures electronic waste as a product, automatically self-balances an object placed on it, and is integrated with Robot Operating System (ROS) for teleoperation using a remote controller.
This allows for ease of control and the range of motion that satisfies our technical requirements. 
To secure the cart on the hoverboard circuit, we mounted the cart using cardboard and tape.
However, the cart height was incompatible with the height of the hoverboard; as a result, we used a metal chassis to lift the cart to promote self-balancing.

This prototype was a unique compact assessment of what could be done and what tasks would take more iterations, allowing acute knowledge for the prototyping process. 
It painted a clear picture of future development directions and sparked discussion among stakeholders about important features for successful deployments in clinical team collaboration. 
We conducted a 1-hour interview with our two medical collaborators to collect feedback on MCCR v1 which involved showing demonstrations of the robot in videos, and asking questions about the robot’s functionality, appearance, and concerns about integrating the robot into medical procedures.
We learned the importance of using a cart that resembles a crash cart and that a mobile cart is useful, but it would be helpful for the robot to provide supply recommendation capabilities.

\subsection{Crash Cart Robot Prototype 2}

After informal validation of the properties of the first MCCR, we proceeded to iterate on the design to accomplish a more holistic prototype that incorporated feedback and lessons from informal suggestions from our medical collaborators. 
In this section, we lay out the design considerations for each of these components. 
A core principle for driving the implementation of this prototype is to keep the robot as close as possible to the original crash cart design. 
The vision for this prototype is to enable the robot to approach healthcare workers during medical procedures and open relevant drawers with equipment to prevent them from shuffling through cart drawers in search for supplies.
We use inspiration from prior cart robots that provide interactive drawer opening capabilities \cite{mok2015place}.
Thus, we focus on developing a modular mobile platform based off a hoverboard, integrating linear actuators for the shelves, and developing Printed Circuit Boards (PCBs) as well as control logic in Robot Operating System (ROS) to control the a robot that resembles drawers of a traditional crash cart with less focus on the appearance of the cart.

We chose a 6-Drawer cart for the second MCCR prototype for several reasons. 
The dimensions are 28" W x 18" D x 34.5" H with a weight capacity of 300 lbs, it has a taller height from the ground than the previous cart, closely resembling a traditional crash cart.
However, we faced challenges finding a red cart of similar dimensions so the cart color is granite.
Lastly, the cart was inexpensive (\$580) compared to a traditional crash cart ($>$\$1000).


Building this prototype involved the design of hardware, software, products using Computer Aided Design (CAD) to connect them as an integrated teleoperated system.
The main design considerations include a Printed Circuit Board (PCB) that controls linear actuators mounted on the cart drawers using a remote controller, 3D printed products to mount the linear actuators on the drawers, and assembling the full integrated circuit on the cart.
All components must fit within the dimensions of the cart and allow for communication between devices over Bluetooth or WiFi.

\begin{figure}[t] 
	\centering 
	\includegraphics[width=1.0\linewidth]{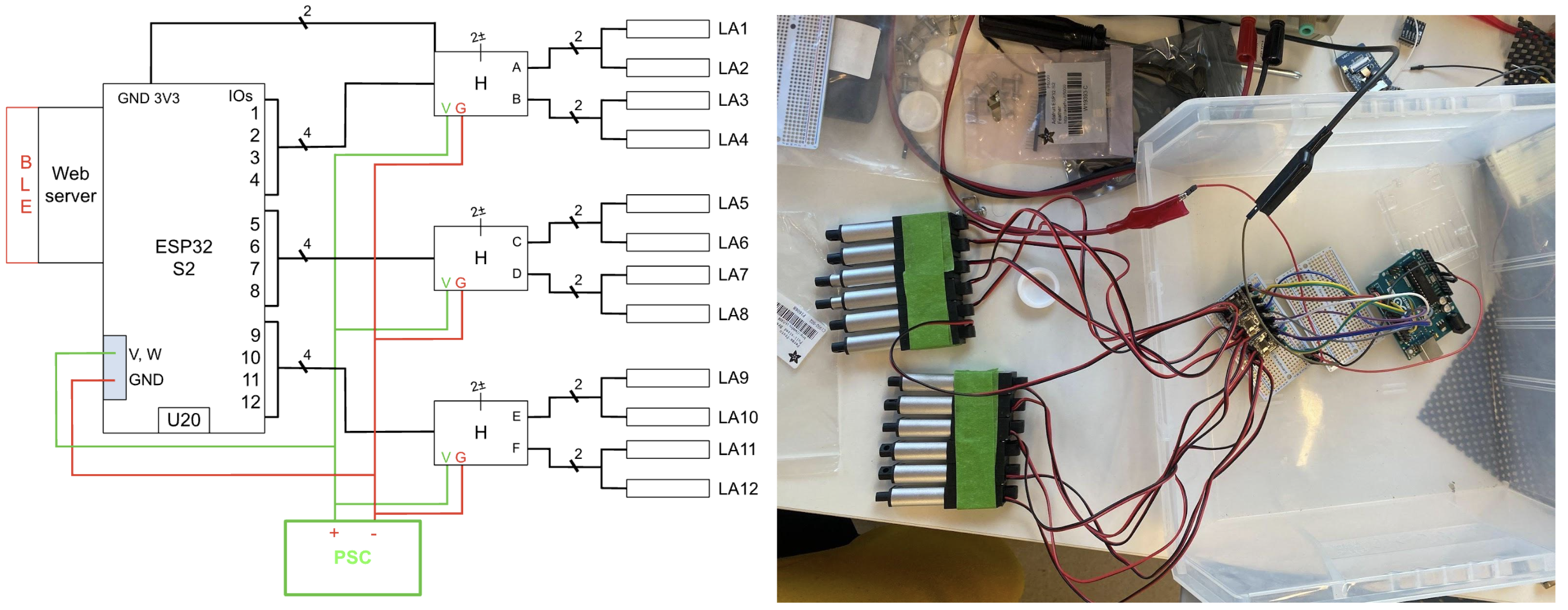} 
	\caption{Schematic of linear actuator array for drawer opening mechanism and CAD of actuator and holder.}
	\label{fig:pcb} 
    \vspace{-5pt}
\end{figure}

\textbf{Printed Circuit Board Design:} The PCB design was an iterative process that involved searching for electronics that enabled the robot to open relevant drawers of the cart, parts that can fit within the spatial constraints of the cart, and power supply requirements that allowed for at least 2 hours of operation (see Figure \ref{fig:pcb}).
This system enables an ER stakeholder to remotely control the MCCR during medical procedures and encourage users to retrieve supplies from relevant drawers. 
The PCB design requirements include connecting two linear actuators to each drawer in parallel, generating enough current to open drawers filled with supplies, as well as opening and closing the linear actuators via predefined wireless buttons using a remote controller over WiFi or Bluetooth.

The most challenging, yet critical, component of the MCCR is the mechanism to open the shelves of the crash cart. 
Requirements include ease of tele-operation, fast response, and constant control (e.g., not pushing out so fast that the tools inside the drawers are lunged outwards into the floor). 
We explored a range of solutions such as spring-based mechanisms with solenoid valves, geared conveyor belts, and magnetic valves. 
To drive the linear actuators, we use basic H-bridges. 
We built a custom PCB containing three H-bridges to control our 12 linear actuators via multiplexing. 
Eventually, linear actuators Mini Electric Linear Actuators were our solution of choice as they best satisfied our mix of requirements including compact solution that fits along the drawers of the cart and has a maximum speech of 1.97 inches per second.

We discovered an effective circuit design through trial-and-error. 
This circuit is controlled with an Arduino UNO microcontroller.
We used an ESP8266 module to enable the microcontroller to connect to 2.4GHz Wi-Fi and three DRV8833 Dual H-Bridge Motor Drivers to open and close 12 linear actuators using a parallel circuit.
We communicate remotely with the Arduino using a paired Bluetooth keyboard over WiFi.
The entire circuit is powered by a 50V battery. 

We conducted experiments to test the speed of the linear actuators by varying the voltage from 5V to 10V to ensure fast movement with out startling users and found 8V provided an appropriate speed. 
Then, we explored how we might position the circuit on the cart, but we found that the Arduino UNO did not fit in cart dimensions.
To address this problem, we used the Adafruit Feather to replace the Arduino UNO because it is smaller in size, compact, and meets our Wi-Fi requirements.
On the more engineering side of the spectrum, this process painted a good picture of the motor strength required to motorize the shelves of the cart of choice and the resulting speed/impulse of movement when the shelf is in motion and how much space it requires to freely move around.

\begin{figure}[t] 
\centering 
\includegraphics[width=1.0\linewidth]{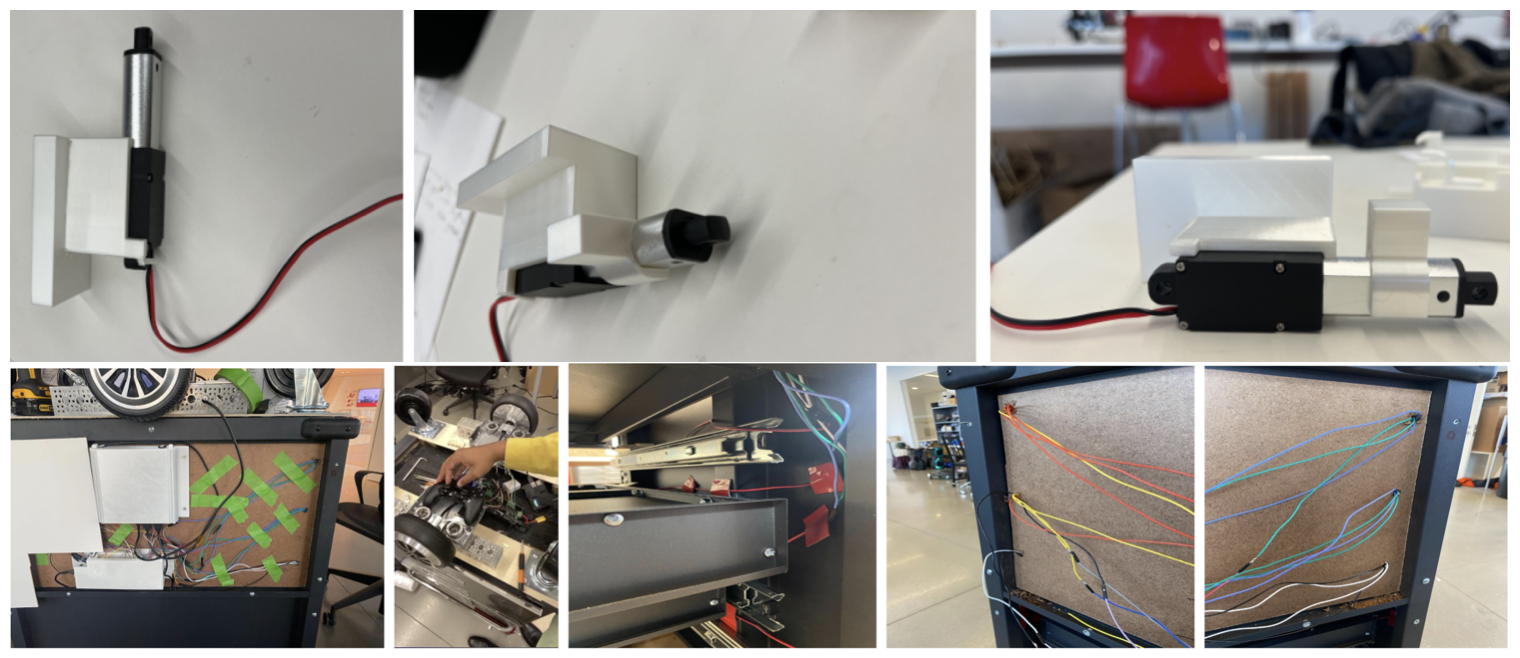} 
\caption{Prototype 2: (Top) The first actuator case prototype pushed the actuator back when it opened so we built a more stable casing to hold it in place. (Bottom) Assembly of hardware components.} 
\vspace{-5pt}
\label{fig:actuator}
\end{figure}

\textbf{Linear Actuator Casings:} To support the actuators, we required a custom 3D printed mount to integrate directly within the MCCR interior walls. 
Given the weight capacity of the shelves (99 lbs), it required a mounting mechanism that rigidly links the actuator to the side wall of the cart.
The actuators are held in two locations, in the back as well as towards the end of the actuator with two mounts as shown in Figure \ref{fig:actuator}.
Each actuator case was created keeping in mind the thickness of the piece and its geometry in order to withstand the recoil of the actuator once the corresponding drawer was pushed open. 
We used Polylactic Acid (PLA) filament as the casing material for simplicity.
After designing the first prototype casing, we found that the actuators would push and slide backwards. 
To address this problem, we designed a blocking wall so the actuator would not push back once activated, effectively holding the actuator in place. 
All products were designed in Fusion 360 and created using a Bambu 3D printer. 

\textbf{Robot Assembly:} To allow others to benefit from our prototype, we describe the stepwise assembly process of building our robot from the individual components which involved mounting the hoverboard and linear actuators, and wiring (see Figure \ref{fig:actuator}). 
To affix the hoverboard \cite{mandel2023recapturing}, the MCCR is equipped with a custom laser-cut acrylic bracket and a modular mount. 
This interface allows for easy adoption to other carts or devices given the ease of laser cutting a new variation. 
It secures a rigid connection with the rest of the crash cart and the weight of the cart on top of it further keeps the mechanism in place.
We drilled holes on the side of the cart to mount the actuators on the cart drawers. 
Then, we wired the circuit to control the actuators between the drawers to the back of the cart along with a 50V battery pack, with at least 2 hours of power supply.

\section{Evaluation}

To better understand the nuances of appropriate robot feedback in acute care settings, we conducted IRB-approved (\#STUDY00008415) field deployments with Wizard-Of-Oz controlled crash cart robots (V2). 
This is a two-day inter-professional medical training event, hosted by a medical school in the global north annually that invites over 130 participants from around the U.S. to participate in mass casualty training sessions, designed by medical educators. 
At this event, medical students engage in training using robotic patient simulators that are teleoperated robots that show physiological signs (e.g., breathing, bleeding, vital signs) to increase the realism of the training scenarios. 
We aimed to explore the medical crash cart robot (V2) that supports teamwork during resuscitation procedures at the patient’s bedside such as delivering supplies or recommending materials. 
The robot was teleoperated by a member of our team to deliver supplies and move the cart toward participants that need to retrieve equipment. 
Participants engaged in a between-subjects study with two conditions: C1) regular crash cart and C2) MCCR.

\subsection{Participants} 

\begin{figure}[t]
    \centering
 \includegraphics[width=0.4\textwidth]{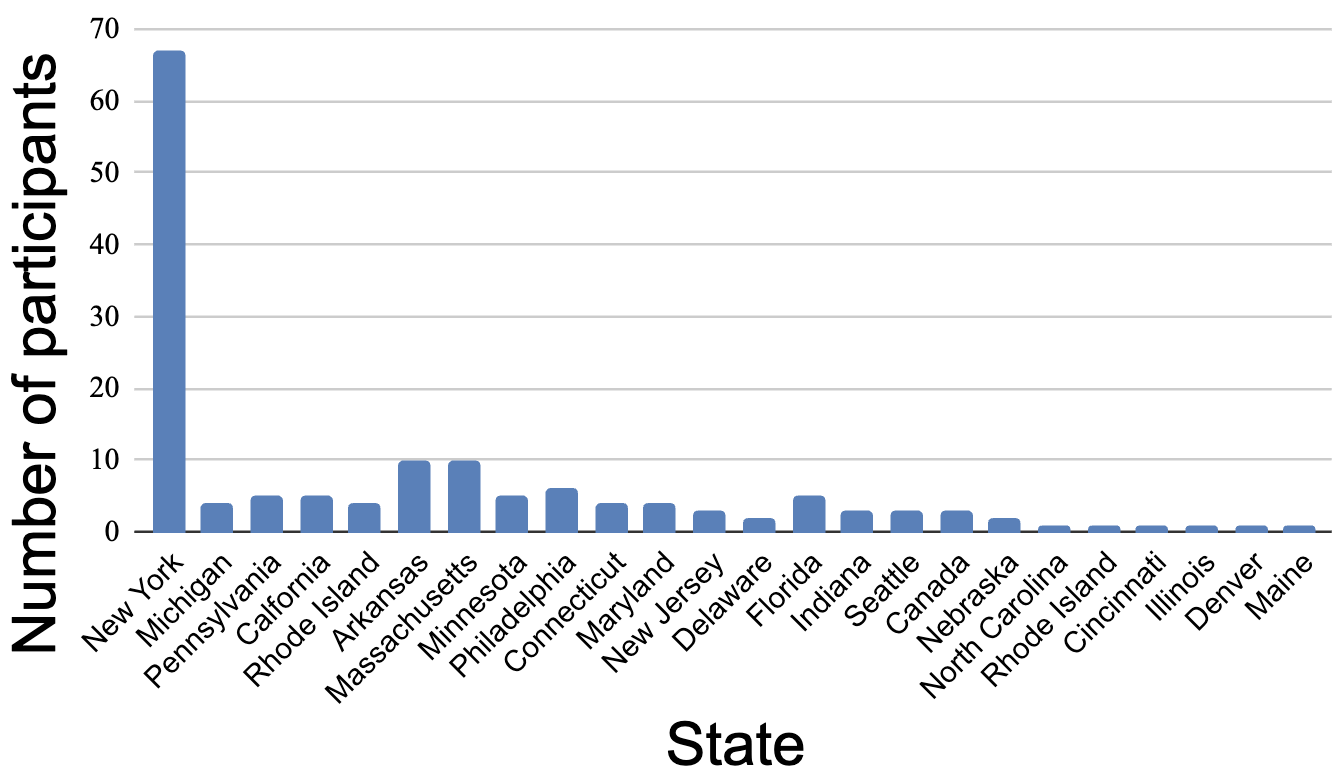}
    \caption{Demographic locations of field deployment attendants.} 
    \label{fig:participants}
\end{figure}

Participants were medical students and practicing healthcare workers from across North America and their expertise included Registered Nursing (36), Child Life Specialists (12), and Fellows (72) from across the global north (see Figure \ref{fig:participants}).
A subset of these participants engaged in 4 experimental groups and 9 control groups with 6-8 people per group.
Due to the nature of field deployment at a public medical training event, we were unable to collect standard demographic information.
Nevertheless, we can confirm that all study participants are over 18 years old.
They traveled from 24 states with most participants from the northeast (67) and 2 countries. 
The majority of participants were females. Specifically, in the four experimental groups, there were 16 females and 5 males. In the nine control groups, it was difficult to determine the exact number of males and females as we only had audio data.

\subsection{Study Task} 

The study tasks involved participants’ performing 30-minute medical training sessions including Joint Teamwork Simulations, Joint Difficult Airway, and Joint Trauma, which were designed by medical educators overseeing the event.
Participants engaged in one of the two conditions.
Next, participants engaged in a 10-minute debrief discussion facilitated by a medical educator to enable participants to discuss what went well, what went wrong, and how they can improve their decision making for the next procedure.
Lastly, we administered post-study surveys to measure workload and usability.

\subsection{Data Collection, Analysis, \& Measures} 

\textbf{Data Recording:} 11 members of our research team assisted with data collection efforts during field deployments. We collected video and audio recordings from 15 45-minute sessions across both experimental and control conditions. For each of these sessions, we captured recordings from two camera angles to support our analysis. Additionally, we administered post-study surveys to both experimental and control groups. We considered it critical to contrast our survey responses and field observations from the intervention condition (robot) against a control. Due to instances of hardware failure, we were unable to capture recordings outside of the 15 sessions mentioned above. In spite of this loss, we retained survey responses from each session including those without video and/or audio data. These recordings, paired with the responses collected from our post-study survey provided a rich source of data for both qualitative and quantitative analysis. 

\textbf{Measures:} We measured participants' workload in terms of demand, physical demand, temporal demand, performance, effort, and frustration using the NASA-TLX scale \cite{hart2006nasa} and MCCR v2 usability using the System Usability Scale \cite{brooke1996sus}.


\textbf{Quantitative Analysis:} To identify statistically significant differences between the study conditions, we ran a t-test to compare the workload of participants. Given the comparatively larger size of the control group, we used the Welch approximation t-test to account for differences in sample sizes, preserving the integrity of the comparison. This approach allowed us to compare the mean scores of the control and robot groups without assuming homogeneity of variances or equal sample sizes, ensuring the comparison was fair and valid.

\textbf{Qualitative Analysis:} We conducted ethnomethodological and conversational analysis (EMCA) \cite{haddington2023emca} on video recordings collected from the study. Our analysis involved multiple steps, followed by a review of the aggregate session data as we documented any instances of interaction between participants and the robot. We categorized interactions as \textit{direct} (i.e., physical human-robot contact, or \textit{indirect} (i.e., emerging from other participants' interactions with the robot, or interactions between the robot and its environment). Throughout this process, we identified common themes of interactions and categorized them into a ``Taxonomy of Failure," focusing on conflicts between robot behavior and user expectations. 

\begin{figure}[t] 
	\centering 
	\includegraphics[width=0.8\linewidth]{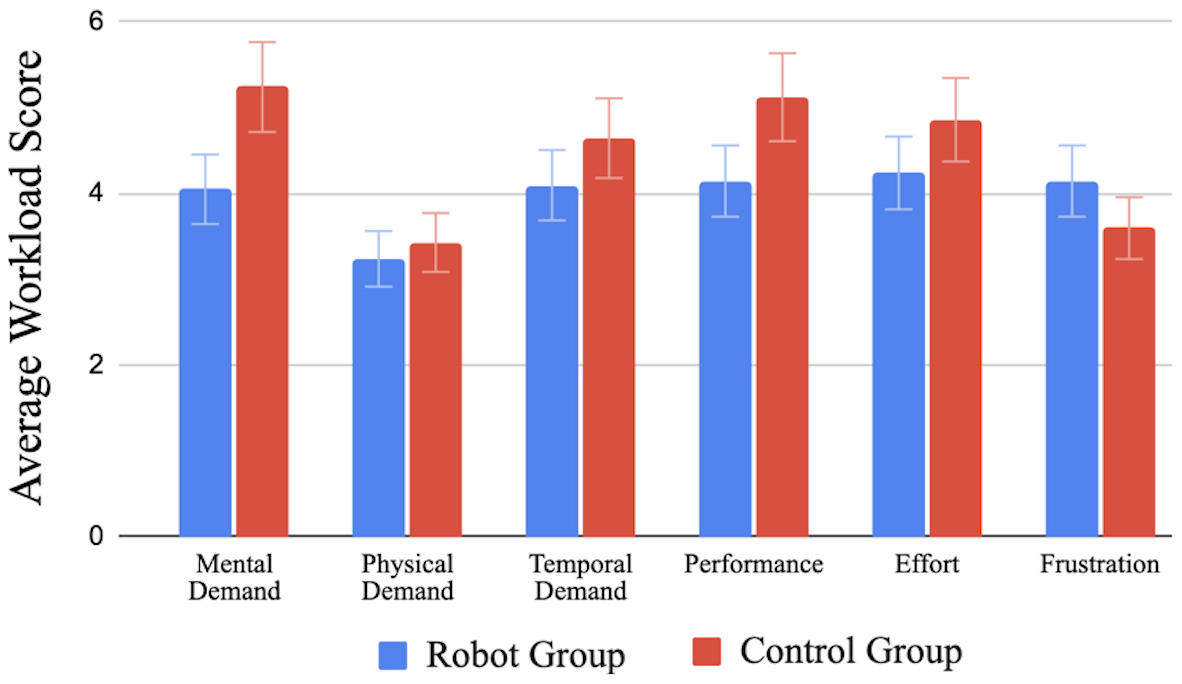} 
	\caption{Medical crash cart robot v2 field deployment results. Control and robot groups NASA-TLX \cite{hart2006nasa} scores.}
	\label{fig:robot-sus-nasa} 
    \vspace{-5pt}
\end{figure}

\section{Quantitative Findings: Workload and Usability}
 \label{sec:quan_results}

Figure \ref{fig:robot-sus-nasa} shows the NASA-TLX results for the experimental and control groups respectively. 
The results indicate that both groups experienced a high workload during the field studies. 
The average perceived workload was slightly higher in the control group (64.0) compared to the Robot Group (56.8). 
After we collected the average NASA-TLX score for each respondent in both groups, a Welch approximation t-test was conducted to assess whether there was a statistically significant difference in workload between the two groups. 
The resulting t-statistic was 2.08 with a p-value of 0.04. 
Given that the p-value is less than the conventional threshold of 0.05, we reject the null hypothesis, indicating that there is a significant difference in workload between the two groups. 
Contrary to what might be expected, the control group reported a significantly higher workload than the robot group, even though both groups were within the ``High Workload" range according to the NASA-TLX scale. 
This suggests that the use of robots in the task may have reduced the perceived workload compared to the traditional control conditions. 

Figure \ref{fig:robot-sus-nasa} shows the results for the MCCR group indicating a SUS score of 39.4 out of 100 (see supplemental material$^1$), in the range of Not Acceptable.
The SUS score indicates that participants are neutral about whether they would use the MCCR. 
This suggests that while the MCCR has utility, participants likely viewed its complexity as a barrier to regular use.
Participants found the MCCR unnecessarily complex and did not find it easy to use, as indicated by the higher score for complexity and a lower score for ease of use. 
This combination suggests that participants experienced difficulties navigating or understanding the MCCR’s functionality.
A significant portion of participants believed they would need the support of a technical person to operate the MCCR effectively. This is a strong indication that the MCCR is not user-friendly or intuitive enough for most participants to handle independently.
While the MCCR’s functions are somewhat well-integrated, there is concern about inconsistencies the robot behavior, which adds to participants’ cognitive load.
Participants expressed that they needed to learn many things before they could get going with the MCCR, which points to a steep learning curve. 
The MCCR was found to be cumbersome to use, and participants reported low confidence while interacting with it.

\section{Qualitative Findings: Taxonomy of Failure}
\label{sec:qual_results}

Our Taxonomy of Failure behavioral conflicts into three categories: (1) Suggestive, (2) Obstructive, and (3) Distractive.

\textbf{Suggestive Failures.} These failures predominantly concern the intuitive—or unintuitive—patterns of interaction facilitated by the cart and their impact on team performance in the emergency room context. 
A key issue observed was the ambiguity of our original signaling system. 
The actuated drawers were activated remotely and served as signals intended to communicate to participants which drawer they should access. 
These signals, however, often failed to convey clear information, leading to confusion about the cart’s behavior. 
For example, our drawer signaling went largely overlooked by a majority of participants. 
Most commonly, this manifested as participants opening several drawers in succession searching for a specific tool or supply. 
In many cases, we found these misalignments of behavior and expectation to complicate and exacerbate confusion during otherwise critical moments.

\textbf{Obstructive Failures.} These failures characterize instances of physical robot movement that modify human behavior. 
Our initial design restricted control of the cart to teleoperation, preventing participants from moving it manually—a limitation stemming from hardware constraints rather than a deliberate design choice. 
This restriction often led to participants' attempting to push the cart manually, only to be met with resistance, disrupting their workflow, and often leading to subsequent distractive failures. 
In some instances, this inability to move the cart by hand resulted in participants having to navigate around it in a tight space, causing delays in reaching essential equipment or positioning. 

\textbf{Distractive Failures.} These failures occur when the cart’s behavior or appearance diverts attention away from primary tasks. 
We observed that the cart’s navigation, while intended to assist in optimizing space and resource management, frequently captured participants’ attention at inopportune moments. 
For example, the cart's sudden movements or unexpected stops, triggered by its navigation and environmental collisions, often drew glances and reactions from participants who were engaged in critical tasks. 
These distractions were compounded by participants’ perception of the cart's novelty and challenges that arose from teleoperation, both further fragmenting the focus of the team. 
In several documented cases, the sudden movement of the cart startled one or more participants, momentarily diverting their attention and prompting verbal exchanges that disrupted the flow of the procedure. 

\section{Prototype 3: Rapid Crash Cart Robotic Feedback System}

\subsection{Approach}

After field deployments, we designed MCCR 3 to improve usability, reduce workload (see Section \ref{sec:quan_results}), and prevent the taxonomy of failures (see Section \ref{sec:quan_results}) using our accumulated findings from all previous rapid prototyping efforts.
A unique benefit of MCCR 3 is its use of multimodal feedback using speech, drawer lights, and alerts to indicate the location of relevant supplies and task reminders.
The design requirements of MCCR 3 include building the robot rapidly with limited complexity to enable others to build a robot with little technical knowledge (see our publically available tutorial$^1$), low-cost, and publically available hardware.
We used 3 primary materials to develop the new MCCR which include a Raspberry Pi, LED light strip, and a  Bluetooth speaker.
We mounted the LED strip along the right/left drawers along the bottom of the cart to the outside of the cart.
We developed a series of Raspberry Pi modules, one for each modality, to enable remote teleoperation of the MCCR.
We connected the speaker and LED light strip to the Raspberry Pi using Bluetooth and wires respectively.
Our modules generate three graphic user interfaces to enable a stakeholder (wizard) to generate appropriate robot feedback on a computer.

To evaluate the MCCR 3, we collected preliminary feedback in informal conversations.
We demonstrated the aformentioned MCCR capabilities in-person to four HCWs with expertise in Emergency Medicine and Clinical Medicine, with 2-26 years of experience, with ages ranging from 28 to 58, and limited knowledge about robots. 
We asked participants to questions such as `How do you envision this system being used in real ERs?’ and `What additional developments and functionalities could be added to make the system effective in real ERs?’
We recorded the conversations in video including audio and images with participants' permission (IRB \#STUDY00008415).
We analyzed the data using grounded theory to identify key themes participants mentioned in their feedback to reflect on the potential benefits and concerns.

\begin{table*}[h]
\caption{Important Design Factors for Crash Cart Robots}
\centering
\footnotesize
\begin{tabular}{|l|l|l|}
\hline
\begin{tabular}[c]{@{}l@{}}\textbf{Communication}: Robots communicate must \\ be clear and quickly provided to users in an \\ intuitive way.\end{tabular}        & \begin{tabular}[c]{@{}l@{}}\textbf{Mobility}: Robotic mobility is more suitable \\ for outside the patient room than inside the \\ patient room.\end{tabular} & \begin{tabular}[c]{@{}l@{}}\textbf{User Adoption}: Robots must provide \\ advanced benefits over traditional crash \\ cart to increase healthcare workers adopt.\end{tabular}     \\ \hline
\begin{tabular}[c]{@{}l@{}}\textbf{Human autonomy}: The robot cannot interrupt \\ human autonomy and must only provide \\ assistance when prompted.\end{tabular} & \begin{tabular}[c]{@{}l@{}}\textbf{Trust}: Robots need to assist healthcare \\ workers accurately and reliably to build \\ user trust over time.\end{tabular} & \begin{tabular}[c]{@{}l@{}}\textbf{Context matters}: Robots should assist \\workers passively in the patient's room\\and deliberately outside the patient's room.
\end{tabular} \\ \hline
\end{tabular}
\label{table:design_factors}
\end{table*}

\subsection{Key Insights}

Our findings highlight important design themes about the MCCR 3's modes of communication during team collaboration.
Two participants found MCCR 3 alerts beneficial for generating reminders to perform repetitive chest compressions using ‘metronome’ sounds. The sounds should be different from standard ER alerts and used to indicate when supply inventory is low.
One participant indicated that the the design of lights outside the MCCR 3 is useful, but it would also be helpful to include LED lights within the drawers to guide HCWs to approximate locations within the drawers (P1) to locate items faster. 
Furthermore, lights tailored to a particular recommended medical task and sequence of items to retrieval could provide more decision-making support (P2). 
Furthermore, participants found dialogue helpful, particularly for standardized procedures and stating the full name of medications to retrieve based on the patient's condition and when the medication should be administered (P2, P4).
Speech and drawer lights can also be used to guide users to a sequence of locations in the cart (e.g., 'Retrieve Epinephrine from drawer 1', then 'Retrieve needles from drawer 2') while activating LEDs for those drawers (P1, P4).

\section{Discussion}

\subsection{Important Crash Cart Robot Design Factors}

Our iterative prototyping process generated important lessons learned and design factors for medical crash cart robots (MCCRs) that assist healthcare workers during medical procedures Emergency Room settings and beyond (see Table \ref{table:design_factors}). 
We found that MCCR mobility can pose unique challenges to medical teams, including frustration and distractions. 
Also, our findings suggest that the use of automated drawer opening for supply recommendations is useful, only when the drawers can automatically close, which presents safety concerns in terms of detecting users' hands to avoid injuries. 
Lastly, we found that MCCRs can create new failures during medical procedures, a setting where human error is a long-standing issue. 
These findings highlight the need for iterative testing and validation of new robotic capabilities in high-stakes environments to ensure that robot failures do not cause additional medical errors.
Recent work in HRI demonstrates to importance of computational models of users' reactions to robot failures \cite{honig2022taxonomy,bremers2023bystander,bremers2023facial}.
However, further research is required to understand how robots can detect when they commit failures, and approaches to enable them to recover from failures in group interactions, which could be particularly useful in healthcare settings to build \textit{trust} with users over time.

Our findings suggest that MCCRs are well-suited to serve as a \textit{reactive actor} in patient rooms to respond to sensor or direct inputs from users and as a \textit{deliberative actor} outside the patient room where the robot assists HCWs to achieve a common goal; in other words, \textit{context matters}. 
As a reactive actor, the MCCR could serve as a \textit{communication} mechanism for error detection when incorrect items are retrieved from the cart by equipping the drawers with motion detection sensors. 
Prior HRI studies relevant to the Emergency Room (ER), focus on telehealth to increase communication between patients and healthcare workers \cite{matsumoto2023robot}, delivery agents that triage patients in the waiting room \cite{wilkes2010heterogeneous}, and receptionist robots \cite{ahn2015healthcare}.
Our study demonstrates new use-cases for MCCRs during ER patient care to provide feedback to HCWs to improve their decision-making in terms of speech, drawer light recommendations, and alerts or sounds. 
More specifically, our study suggests that equipping the robot with lights in the cart drawers could guide users to retrieve relevant items for patient care. 
These capabilities combined could serve as an inventory tracking system that passively notifies users when supplies are in low supply. 
For instance, active robot drawer lights would only be visible to users actively using the cart, speech can indicate a sense of urgency, only speaking to the users when a patient safety risk has been identified by the robot. 

Another important lesson 
is that MCCRs can introduce safety risks. 
During our studies, we found that participants became frustrated when the robot moved toward a user who appeared to be approaching the cart. 
For example, one participant yelled at the robot in frustration during field deployments because it moved while this user was attempting to retrieve items from the cart. 
This caused additional distractions and could negatively impact user trust in the \textit{robot}. 
Thus, healthcare workers' \textit{autonomy} must be preserved to allow them to override the robot when needed, such as stopping it to avoid distractions. This need for worker autonomy was observed not only in healthcare settings \cite{taylor2024towards} but also in other contexts \cite{lee2024contrasting}.

Participants reflected on potential robotic use cases after field studies, particularly those that involve tasks inside and outside the patient's room. 
For example, the MCCR could indicate when inventory is low, navigate to a supply room, and request HCWs' help to restock the cart during downtime. 
Moreover, through inventory tracking, the MCCR could support documentation efforts after procedures to indicate what supplies are retrieved to be entered into the electronic healthcare record. 
Thus, through inventory tracking, the MCCR could provide insights into approximate time-periods of specific medical tasks to help healthcare workers recall when these tasks were performed, which is particularly useful for the recorder in resuscitation procedures. 
Our findings are consistent with prior research on HRI in action teams which demonstrates the need for passive and deliberative (or proactive) MCCR behavior \cite{jamshad2024taking} that adapts to the team in a manner that avoids interruptions during time-critical tasks.

\subsection{Limitations and Future Work}

Our study has some limitations.
Participants in field studies are limited to Pediatric workers with Child Life Specialists, Registered Nurses, M.D. Fellows specialties and they performed three medical scenarios; thus, our findings may not be generalizable to care for adults or those with other expertise.
MCCR 3 was not tested with users; thus, our findings do not reflect the usefulness of speech, alerts/sounds, and drawer lights for recommendations of supplies, error detection, and procedural steps during patient care.

We plan to address these limitations in future work.
Recently, we set up a patient room experimental testbed in our lab to conduct in-lab studies to test future versions of the robot with stakeholders of different expertise. 
Furthermore, we plan to explore more medical training scenarios in collaboration with our medical collaborators to build robots that can work across multiple medical scenarios.

We hope this research inspires HRI researchers to explore robot design for high-stakes environments. 
While we realize this approach to HRI research is time-consuming and requires iterative user engagement, it is a worthwhile endeavor to improve the experiences of people in real-world settings. 

\section{Acknowledgments}

This material was supported by the National Science Foundation under Grant No.
IIS-2423127. We thank Amazia Thompson, Anaiya Badi, and Huajie Cao for their contributions to this research.


\bibliographystyle{IEEEtran}
\bibliography{references}

\end{document}